\documentclass[preprint,5p,twocolumn]{elsarticle}

\usepackage{amssymb}
\usepackage{amsmath}
\DeclareMathOperator*{\argmax}{argmax}

\usepackage{graphicx}
\graphicspath{{images/}}

\usepackage{subcaption}

\usepackage[]{algorithm2e}

\usepackage{hyperref}

\journal{nowhere yet!}

\begin{document}


\newcommand{\myfigure}[3] {
    \begin{figure}
        \centering
        \includegraphics[width=1.0\columnwidth]{#1}
        \caption{#2}
        \label{#3}
    \end{figure}
}

\newcommand{\vect}[1]{\boldsymbol{#1}}

\begin{frontmatter}


\title{Condensed Representation of Machine Learning Data}

\author[1,3]{Rahman Salim Zengin
\corref{cor1}
\fnref{fn1}}
\ead{rszengin@itu.edu.tr}

\author[2,3]{Volkan Sezer
\fnref{fn2}}
\ead{sezerv@itu.edu.tr}

\cortext[cor1]{Corresponding author}

\fntext[fn1]{\href{https://orcid.org/0000-0002-3104-4677}{ORCID: https://orcid.org/0000-0002-3104-4677}}

\fntext[fn2]{\href{https://orcid.org/0000-0001-9658-2153}{ORCID: https://orcid.org/0000-0001-9658-2153}}

\address[1]{Department of Mechatronics Engineering, Istanbul Technical University, Istanbul, Turkey}
\address[2]{Department of Control and Automation Engineering, Istanbul Technical University, Istanbul, Turkey}
\address[3]{Autonomous Mobility Group, Electrical and Electronics Engineering Faculty, Istanbul Technical University, Istanbul, Turkey}

\begin{abstract}
    Training of a Machine Learning model requires sufficient data. The
    sufficiency of the data is not always about the quantity, but about the
    relevancy and reduced redundancy. Data-generating processes create massive
    amounts of data. When used raw, such big data is causing much computational
    resource utilization. Instead of using the raw data, a proper Condensed
    Representation can be used instead. Combining K-means, a well-known
    clustering method, with some correction and refinement facilities a novel
    Condensed Representation method for Machine Learning applications is
    introduced. To present the novel method meaningfully and visually,
    synthetically generated data is employed. It has been shown that by using
    the condensed representation, instead of the raw data, acceptably accurate
    model training is possible.
\end{abstract}


\end{frontmatter}

\section{Introduction}
\label{sec:introduction}

The concept of condensed representation is considered in
\cite{mannila_data_1996}, \cite{mannila_methods_1997} as determining frequent
patterns in big databases. It is given as an idea to improve pattern matching
against big databases. Instead of using the actual database, a condensed
representation can be queried for efficiency. The condensed representation
should give approximately correct results.

Online probability density estimation is used in
\cite{geilke_probabilistic_2014} as a method of condensed representation for
data mining tasks, especially when not having the whole data envisioned. 

In \cite{geilke_modeling_2015}, online density estimators are employed to obtain
condensed representations from data streams.

Wavelet-based PDFs are used in \cite{peel_distinguishing_2019} for condensed
representation of training data to train a CNN. This usage resulted in less
computational resource usage.

The resources in the literature, such as
\cite{soulet_condensed_2004,adams_condensed_2009,buntine_condensed_2009}, are
mainly related but not limited to, data mining and set operations on sequential
datasets.

The most relevant machine learning subject to the study of this paper is
autoencoders \cite{skansi_autoencoders_2018}. Autoencoders represent data in a
new space with different dimensionality.

It has been shown that autoencoders can generate condensed representations of
the sequential input data \cite{berthold_dual_2020}. Then a classifier can
process the representations instead of the original data.

Autoencoders are used in \cite{kempinska_modelling_2019} to obtain condensed
representations of the urban road networks.

The main goal of this paper is to introduce a novel approach to the condensed
representation of machine learning data and to demonstrate its practical use.
Accordingly, the principal components of the novel method are defined, combined
and tested on some pretty standart synthetic data. Then, the final thoughts are
shared. 

\section{The Algorithm}
\label{sec:the_algorithm}

The algorithm consists of two main steps. In the initialization step, initial
representation centers are placed over the data. Following the initialization,
the refinement process improves the purities of the classes.

\subsection{Initialization}
\label{sub:initialization}

The initialization process starts with the
separation of different classes of data instances. Given labels of the dataset
are used to discriminate classes. Then, to distribute initial Condensed
Representation Centers (CRCs) over the data, the k-Means algorithm is applied to
every class of the data separately. The k parameter can be chosen independently
for every class.

The resulting cluster centers, the initial CRCs, are placed over the whole
dataset. Every data instance is assigned to the nearest CRC. So, the space is
tessellated into convex polyhedral cells, which the enclosed CRC defines.

\begin{figure*}
    \centering
    \begin{subfigure}[b]{0.32\textwidth}
        \centering
        \includegraphics[width=\textwidth]{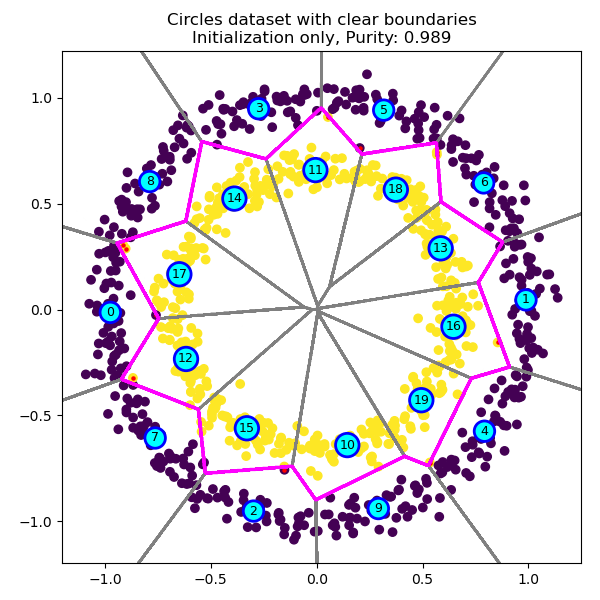}
        \caption{Clear boundaries, initialized}
        \label{fig:circles_init:1}
    \end{subfigure}
    \begin{subfigure}[b]{0.32\textwidth}
        \centering
        \includegraphics[width=\textwidth]{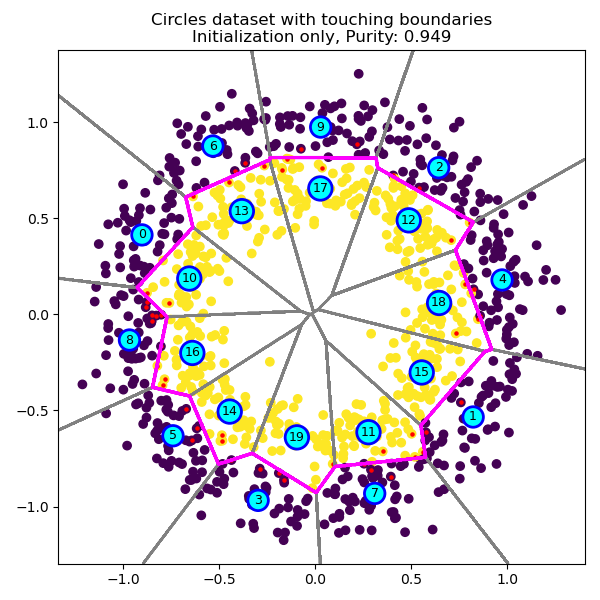}
        \caption{Touching boundaries, initialized}
        \label{fig:circles_init:2}
    \end{subfigure}
    \begin{subfigure}[b]{0.32\textwidth}
        \centering
        \includegraphics[width=\textwidth]{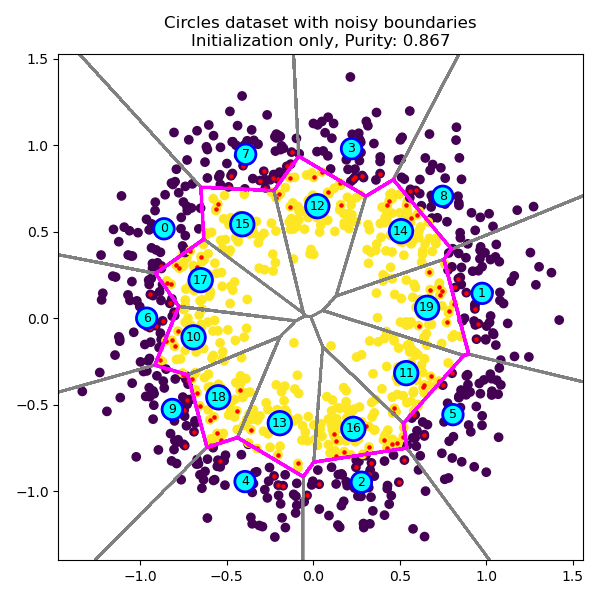}
        \caption{Noisy boundaries, initialized}
        \label{fig:circles_init:3}
    \end{subfigure}
    \caption{Condensed representations of synthetic circles dataset, initialization only}
    \label{fig:circles_init}

    \vspace{0.5cm}

    \begin{subfigure}[b]{0.32\textwidth}
        \centering
        \includegraphics[width=\textwidth]{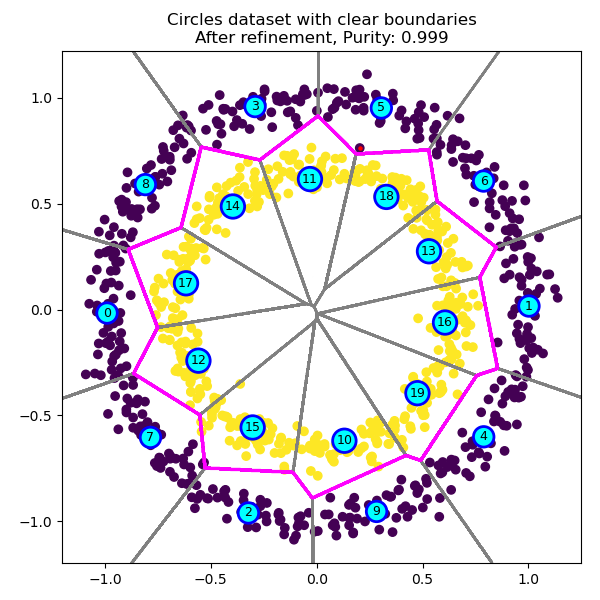}
        \caption{Clear boundaries, refined}
        \label{fig:circles_refine:1}
    \end{subfigure}
    \begin{subfigure}[b]{0.32\textwidth}
        \centering
        \includegraphics[width=\textwidth]{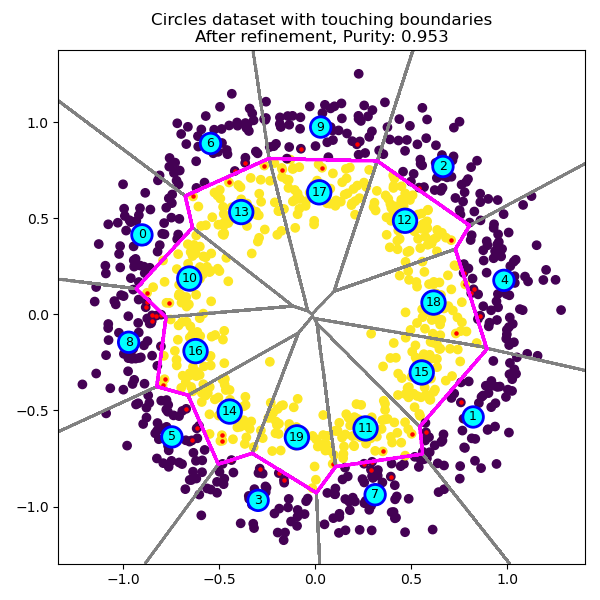}
        \caption{Touching boundaries, refined}
        \label{fig:circles_refine:2}
    \end{subfigure}
    \begin{subfigure}[b]{0.32\textwidth}
        \centering
        \includegraphics[width=\textwidth]{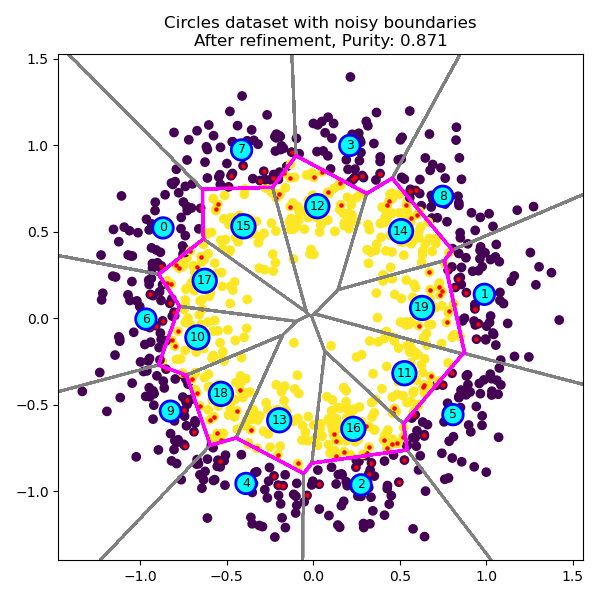}
        \caption{Noisy boundaries, refined}
        \label{fig:circles_refine:3}
    \end{subfigure}
    \caption{Condensed representations of synthetic circles dataset, after refinement}
    \label{fig:circles_refine}
\end{figure*}

\begin{figure*}
    \centering
    \begin{subfigure}[b]{0.32\textwidth}
        \centering
        \includegraphics[width=\textwidth]{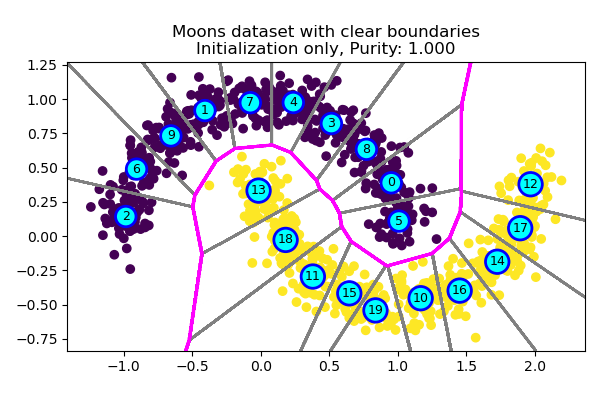}
        \caption{Clear boundaries, initialized}
        \label{fig:moons_init:1}
    \end{subfigure}
    \begin{subfigure}[b]{0.32\textwidth}
        \centering
        \includegraphics[width=\textwidth]{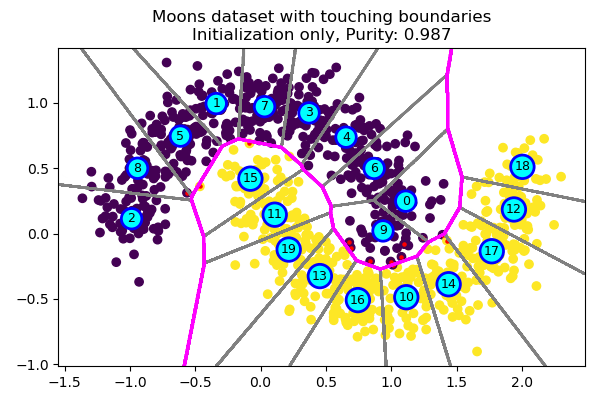}
        \caption{Touching boundaries, initialized}
        \label{fig:moons_init:2}
    \end{subfigure}
    \begin{subfigure}[b]{0.32\textwidth}
        \centering
        \includegraphics[width=\textwidth]{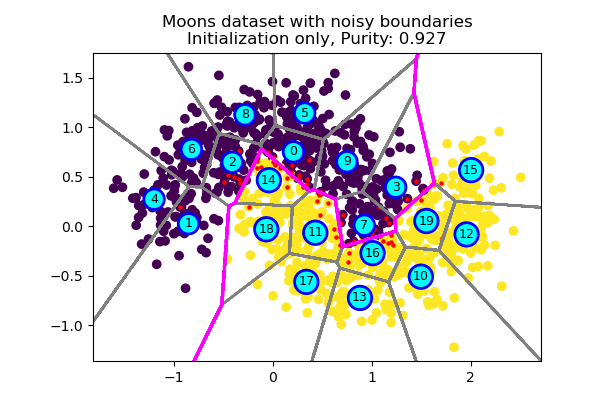}
        \caption{Noisy boundaries, initialized}
        \label{fig:moons_init:3}
    \end{subfigure}
    \caption{Condensed representations of synthetic moons dataset, initialization only}
    \label{fig:moons_init}

    \vspace{0.5cm}

    \begin{subfigure}[b]{0.32\textwidth}
        \centering
        \includegraphics[width=\textwidth]{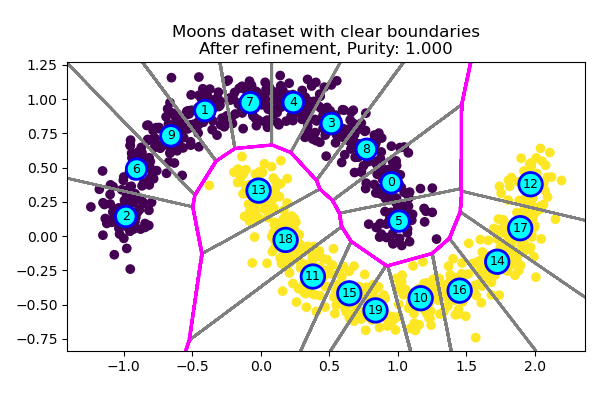}
        \caption{Clear boundaries, refined}
        \label{fig:moons_refine:1}
    \end{subfigure}
    \begin{subfigure}[b]{0.32\textwidth}
        \centering
        \includegraphics[width=\textwidth]{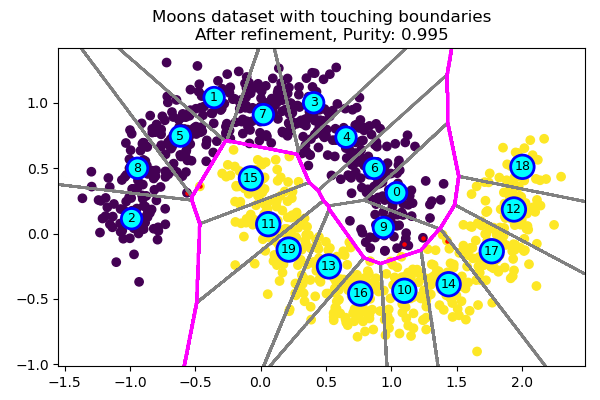}
        \caption{Touching boundaries, refined}
        \label{fig:moons_refine:2}
    \end{subfigure}
    \begin{subfigure}[b]{0.32\textwidth}
        \centering
        \includegraphics[width=\textwidth]{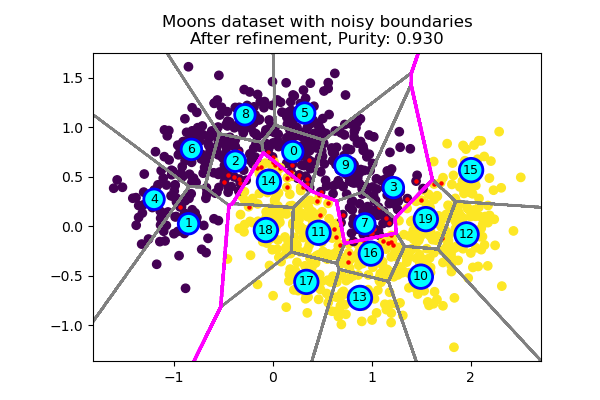}
        \caption{Noisy boundaries, refined}
        \label{fig:moons_refine:3}
    \end{subfigure}
    \caption{Condensed representations of synthetic moons dataset, after refinement}
    \label{fig:moons_refine} 
\end{figure*}

When all classes are combined, some of the CRCs might lie over the intersections
of different classes. In order to determine the represented class of a CRC,
plurality voting is applied to the covered data instances. The result of the
vote becomes labels of the CRCs. (Algorithm \ref{alg:initialization}) (Figure
\ref{fig:circles_init}) (Figure \ref{fig:moons_init})

\begin{algorithm}
    \DontPrintSemicolon
    \SetKwProg{Fn}{Function}{}{}
    \SetKwFunction{Init}{Initialize}
    \Fn(){\Init{$n_{centers}$}}{}{
        \KwData{\\
            $X \longleftarrow \begin{bmatrix}
                \vect{x_1} & \cdots & \vect{x_N}
            \end{bmatrix}$ \tcc{Instances}
            $y \longleftarrow \begin{bmatrix}
                y_1 & \cdots & y_N
            \end{bmatrix}$ \tcc{Labels}
        }
        \KwResult{$Centers$ \tcc*[h]{Condensed Representation Centers}}
        \Begin{
            \For(\tcc*[h]{every class}){$i = {1, \ldots, n_{classes}}$}{
                $Centers[i] \longleftarrow$ KMeans($X[y = i]$, $k_i$)
            }
        }
    }
    \caption{Initialization}
    \label{alg:initialization}
\end{algorithm}

\subsection{Refinement}
\label{sub:refinement}

The refinement process advances CRCs to improve the purities of the enclosed
cells. Several suboperations are required to accomplish the refinement process.
The base elements of the refinement should be defined before explaining the
overall process. (Figure \ref{fig:circles_refine}) (Figure
\ref{fig:moons_refine})

\subsubsection{Labeling}
\label{subsub:labeling}

Some of the CRCs might lie over the intersections of different classes. In order
to determine the represented classes of the CRCs, plurality voting is applied to
the assigned data instances. The result of the vote becomes labels of the CRCs.
(Algorithm \ref{alg:labeling})

\begin{algorithm}
    \DontPrintSemicolon
    \SetKwProg{Fn}{Function}{}{}
    \SetKwFunction{Label}{Labeling}
    \Fn(){\Label{}}{}{
        \KwData{\\
            $X \longleftarrow \begin{bmatrix}
                \vect{x_1} & \cdots & \vect{x_N}
            \end{bmatrix}$ \tcc{Data Instances}
            $y \longleftarrow \begin{bmatrix}
                y_1 & \cdots & y_N
            \end{bmatrix}$ \tcc{Data Labels}
            $Centers \longleftarrow \begin{bmatrix}
                \vect{c_1} & \cdots & \vect{c_{n_{centers}}}
            \end{bmatrix}$ \tcc{CRCs}
        }
        \KwResult{$Labels$ \tcc*[h]{Labels of the centers}}
        \Begin{
            $Assignments \longleftarrow$ Assign($X, Centers$)\;
            \For(\tcc*[h]{every center}){$i = {1, \ldots, n_{centers}}$}{
                $Classes, Counts \longleftarrow$ CountUnique($y[Assignments = i]$)
                $Labels[i] \longleftarrow Classes[\argmax Counts]$ 
            }
        }
    }
    \caption{Labeling}
    \label{alg:labeling}
\end{algorithm}

\subsubsection{Purity Metric}
\label{subsub:purity}

The purity of data assignments to CRCs is measured to determine
how clearly class boundaries are set. Assignment purities are measured
independently for every CRC. This metric is used to guide the convergence of the
refinement. (Algorithm \ref{alg:purities})

\begin{algorithm}
    \DontPrintSemicolon
    \SetKwProg{Fn}{Function}{}{}
    \SetKwFunction{Pure}{Purities}
    \Fn(){\Pure{}}{}{
        \KwData{\\
            $X \longleftarrow \begin{bmatrix}
                \vect{x_1} & \cdots & \vect{x_N}
            \end{bmatrix}$ \tcc{Data Instances}
            $y \longleftarrow \begin{bmatrix}
                y_1 & \cdots & y_N
            \end{bmatrix}$ \tcc{Data Labels}
            $Centers \longleftarrow \begin{bmatrix}
                \vect{c_1} & \cdots & \vect{c_{n_{centers}}}
            \end{bmatrix}$ \tcc{Condensed Representation Centers}
        }
        \KwResult{$Purities$ \tcc*[h]{Purities of the assignments}}
        \Begin{
            $Assignments \longleftarrow$ Assign($X, Centers$)\;
            \For(\tcc*[h]{every center}){$i = {1, \ldots, n_{centers}}$}{
                $Counts \longleftarrow$ CountUnique($y[Assignments = i]]$)
                $Purities[i] \longleftarrow \max Counts \div \sum Counts$ 
            }
        }
    }
    \caption{Purities}
    \label{alg:purities}
\end{algorithm}

\subsubsection{Population Counts}
\label{subsub:population}

The number of data instances under the containment of a CRC specifies the
importance and effectiveness within the total coverage. Less populated
polyhedral cells affect the complete purity lesser. (Algorithm
\ref{alg:popcount})

\begin{algorithm}
    \DontPrintSemicolon
    \SetKwProg{Fn}{Function}{}{}
    \SetKwFunction{Popcnt}{Popcount}
    \Fn(){\Popcnt{}}{}{
        \KwData{\\
            $X \longleftarrow \begin{bmatrix}
                \vect{x_1} & \cdots & \vect{x_N}
            \end{bmatrix}$ \tcc{Data Instances}
            $Centers \longleftarrow \begin{bmatrix}
                \vect{c_1} & \cdots & \vect{c_{n_{centers}}}
            \end{bmatrix}$ \tcc{CRCs}
        }
        \KwResult{\\
        $IDs$ \tcc*[h]{Identifiers of the centers}\;
        $Counts$ \tcc*[h]{Population counts}}
        \Begin{
            $Assignments \longleftarrow$ Assign($X, Centers$)\;
            $IDs,Counts \longleftarrow$ CountUnique($Assignments$)
        }
    }
    \caption{Population counts}
    \label{alg:popcount}
\end{algorithm}

\subsubsection{Soft Assignment}
\label{subsub:soft_assignment}

Although for the purity metric, absolute assignments of the data instances to
the CRCs are considered, for the refinement, a weighted soft assignment scheme
is used. A data instance is partially assigned to the nearest and the second
nearest CRCs. Assignment weights for both CRCs are calculated by applying a
weighting function. The assignment weights are equal if the distances are equal.
As one CRC gets closer, the relevant assignment weight approaches unity.
(Algorithm \ref{alg:softassign})

\begin{algorithm}
    \DontPrintSemicolon
    \SetKwProg{Fn}{Function}{}{}
    \SetKwFunction{SoftAssign}{SoftAssign}
    \Fn(){\SoftAssign{}}{}{
        \KwData{\\
            $X \longleftarrow \begin{bmatrix}
                \vect{x_1} & \cdots & \vect{x_N}
            \end{bmatrix}$ \tcc{Data Instances}
            $Centers \longleftarrow \begin{bmatrix}
                \vect{c_1} & \cdots & \vect{c_{n_{centers}}}
            \end{bmatrix}$ \tcc{CRCs}
        }
        \KwResult{\\
        $IDs$ \tcc*[h]{Identifiers of the centers}\;
        $Weights$ \tcc*[h]{Assignment weights}}
        \Begin{
            $IDs,Distances \longleftarrow$ TwoNearest($X, Centers$)
            $Weights \longleftarrow 1 - \frac{Distances^2}{\sum Distances^2}$\;
        }
    }
    \caption{Soft assignment}
    \label{alg:softassign}
\end{algorithm}

\subsubsection{Correctness}
\label{subsub:correctness}

If a data instance is assigned to a CRC, for a correct assignment, the label of
that CRC must be equal to the given label of the data. In the soft assignment
case, both CRCs' labels are considered. The correctnesses of those comparisons
is specific to the current CRC configurations and might change with a refinement
iteration. (Algorithm \ref{alg:correctness})

\begin{algorithm}
    \DontPrintSemicolon
    \SetKwProg{Fn}{Function}{}{}
    \SetKwFunction{Correctness}{Correctness}
    \Fn(){\Correctness{}}{}{
        \KwData{\\
            $Labels$ \tcc*[h]{Labels of the centers}\;
            $IDs$ \tcc*[h]{Soft assignment identifiers}\;
            $y \longleftarrow \begin{bmatrix}
                y_1 & \cdots & y_N
            \end{bmatrix}$ \tcc{Data Labels}
        }
        \KwResult{\\
        $Correctness$ \tcc*[h]{Correctnesses of the soft assignments}}
        \Begin{
            $Correctness \longleftarrow Labels[IDs] \equiv y$\;
        }
    }
    \caption{Correctness}
    \label{alg:correctness}
\end{algorithm}

\subsubsection{Activity Group}
\label{subsub:activity}

A refinement iteration is not executed over the whole dataset. The activity
group of the refinement is determined by checking the correctnesses of the first
(nearest) and the next (second nearest) CRCs. If both correctnesses are true,
the data instance is under a proper cover and needs no more intervention. If
both are false, the data instance can be treated as an outlier where it stays.

A data instance is accepted within the activity group only if its soft
assignment is shared between a true and a false CRC. (Eq \ref{eq:activity})

\begin{equation}
    Activity = FirstCorrectness \neq SecondCorrectness
    \label{eq:activity}
\end{equation}

\subsubsection{Advancement Vectors}
\label{subsub:advancement_vectors}

True assignments attract CRCs, and false assignments repel. For a CRC, when the
effects of all assignments are combined, it becomes the advancement vector of
that CRC. The advancement vectors are calculated using weighted vector
differences between assigned data instances and CRCs. Weights come from soft
assignments and are signed according to the correctnesses. (Algorithm
\ref{alg:advancement_vectors})

\begin{algorithm}
    \DontPrintSemicolon
    \SetKwProg{Fn}{Function}{}{}
    \SetKwFunction{AdvanVect}{AdvanVect}
    \Fn(){\AdvanVect{}}{}{
        \KwData{\\
            $X \longleftarrow \begin{bmatrix}
                \vect{x_1} & \cdots & \vect{x_N}
            \end{bmatrix}$ \tcc{Data Instances}
            $Centers \longleftarrow \begin{bmatrix}
                \vect{c_1} & \cdots & \vect{c_{n_{centers}}}
            \end{bmatrix}$ \tcc{CRCs}
        }
        \KwResult{\\
            $A \longleftarrow \begin{bmatrix}
            \vect{a_1} & \cdots & \vect{a_{n_{centers}}}
            \end{bmatrix}$ \tcc{Adv vecs}}
        \Begin{
            \For(\tcc*[h]{every center}){$i = {1, \ldots, n_{centers}}$}{
                \tcc{Diff assigned instances with the current center and multiply with signed weights}
                $a_i \longleftarrow ((X_i - c_i) \times Weights_i).mean()$
            }
        }
    }
    \caption{Advancement vectors}
    \label{alg:advancement_vectors}
\end{algorithm}

\subsubsection{Selective Advancement}
\label{subsub:selective_advancement}

Advancement of CRCs is selectively applied depending on the purity improvements.
If the updated state of a CRC results in better purity, the up-to-date state is
used; otherwise, the previous state is kept. (Algorithm
\ref{alg:selective_advancement})

\begin{algorithm}
    \DontPrintSemicolon
    \SetKwProg{Fn}{Function}{}{}
    \SetKwFunction{SelectAdvance}{SelectAdvance}
    \Fn(){\SelectAdvance{}}{}{
        \KwData{\\
            $A \longleftarrow \begin{bmatrix}
                \vect{a_1} & \cdots & \vect{a_{n_{centers}}}
                \end{bmatrix}$ \tcc{Advancement vectors}
            $Purity_{before}$ \tcc*[h]{Purities before}\;
            $Purity_{after}$ \tcc*[h]{Purities after}\;
        }
        \KwResult{\\
            $A_{selected} \longleftarrow \begin{bmatrix}
            \vect{a_1} & \cdots & \vect{a_{n_{centers}}}
            \end{bmatrix}$ \tcc{Selected advancement vectors}
        }
        \Begin{
            \For(\tcc*[h]{every center}){$i = {1, \ldots, n_{centers}}$}{
                \eIf(\tcc*[h]{Similar or worse purity}){$Purity_{before}^{[i]} \ge Purity_{after}^{[i]}$}{
                    $A_{selected}^{[i]} \longleftarrow Purity_{before}^{[i]}$\;
                }(\tcc*[h]{Better purity}){
                    $A_{selected}^{[i]} \longleftarrow Purity_{after}^{[i]}$\;
                }
            }
        }
    }
    \caption{Selective advancement}
    \label{alg:selective_advancement}
\end{algorithm}

\subsubsection{Convergence}
\label{subsub:convergence}

During the refinement, the general purity of the present CRCs configuration is
tracked for every iteration. Parameters of the purest configuration are saved.
If the purity starts to go down, the refinement process is ended. (Algorithm
\ref{alg:convergence})
 
\begin{algorithm}
    \DontPrintSemicolon
    \SetKwProg{Fn}{Function}{}{}
    \SetKwFunction{Convergence}{Convergence}
    \Fn(){\Convergence{}}{}{
        \KwData{\\
            $OverallPurity$ \tcc*[h]{The overall purity}\;
            $ModelParams$ \tcc*[h]{The current parameters}\;
        }
        \KwResult{\\
            $BestPurity$ \tcc*[h]{The best overall purity}\;
            $BestParams$ \tcc*[h]{The best parameters}\;
        }
        \Begin{
            $BestPurity \longleftarrow OverallPurity$\;
            \While(\tcc*[h]{Until the best overall purity is reached}){$OverallPurity \ge BestPurity$}{
                $BestPurity \longleftarrow OverallPurity$\;
                $BestParams \longleftarrow ModelParams$\;
                \tcc*[h]{Do the refinement}\;
                Refine()
            }
        }
    }
    \caption{Convergence}
    \label{alg:convergence}
\end{algorithm}

\subsubsection{The Overall Refinement Process}
\label{subsub:overall_refinement}

In every iteration, soft assignment is applied to the whole dataset for the
current state of the CRCs. Then, the correctnesses of the primary and secondary
assignments are determined. Afterward, the current activity group is designated
using the correctnesses of the soft assignments. Later, the advancement
candidate of every CRC is found considering weighted and polarized advancement
vectors. Finally, selective advancement is applied, and the next state is
reached. (Algorithm \ref{alg:refine})

\begin{algorithm}
    \DontPrintSemicolon
    \SetKwProg{Fn}{Function}{}{}
    \SetKwFunction{Refine}{Refine}
    \Fn(){\Refine{}}{}{
        \KwData{\\
            $X$ \tcc*[h]{Data Instances}\;
            $y$ \tcc*[h]{Data Labels}\;
            $Centers$ \tcc*[h]{CRCs}\;
            $Labels$ \tcc*[h]{Labels of the centers}\;
        }
        \KwResult{\\
            $NewCenters$ \tcc*[h]{New CRCs}\;
            $NewLabels$ \tcc*[h]{New labels of the centers}\;
        }
        \Begin{
            $IDs, Weights \longleftarrow$ SoftAssign($X, Centers$)\;
            $A \longleftarrow$ AdvanVect($X, Centers$)\;
            $A_{selected} \longleftarrow$ SelectAdvance($A, Purity_{before}, Purity_{after}$)\;
            $NewCenters \longleftarrow Centers + A_{selected}$\;
            $NewLabels \longleftarrow$ Labeling($X, y, NewCenters$)\;
        }
    }
    \caption{Refinement}
    \label{alg:refine}
\end{algorithm}

\section{Experimental Results}
\label{sec:experimental}

The platform used for the experimental tests and comparisons is as follows: The
CPU used for the experimentation is Intel(R) Core(TM) i7-7700 running at 3.60GHz
frequency; the system has 32GB of DDR4 RAM running at 2133 MHz. Nvidia Geforce
GTX 1080Ti is the system GPU.

\begin{figure*}
    \centering
    \begin{subfigure}[b]{0.32\textwidth}
        \centering
        \includegraphics[width=\textwidth]{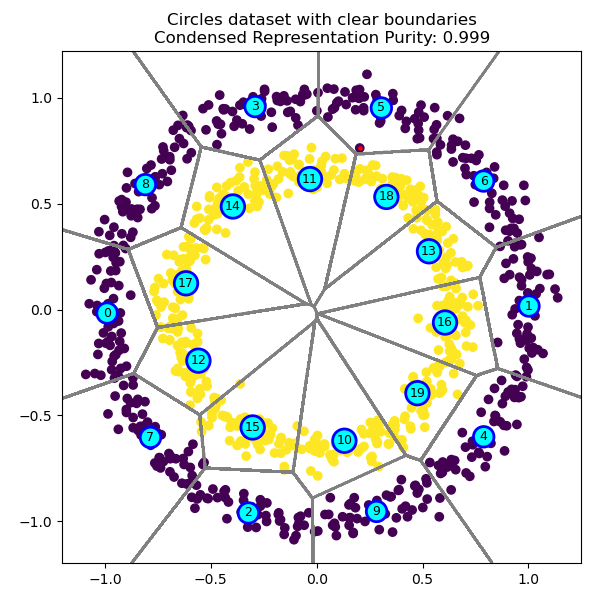}
        \caption{Clear boundaries, condensed}
        \label{fig:test_circles:1}
    \end{subfigure}
    \begin{subfigure}[b]{0.32\textwidth}
        \centering
        \includegraphics[width=\textwidth]{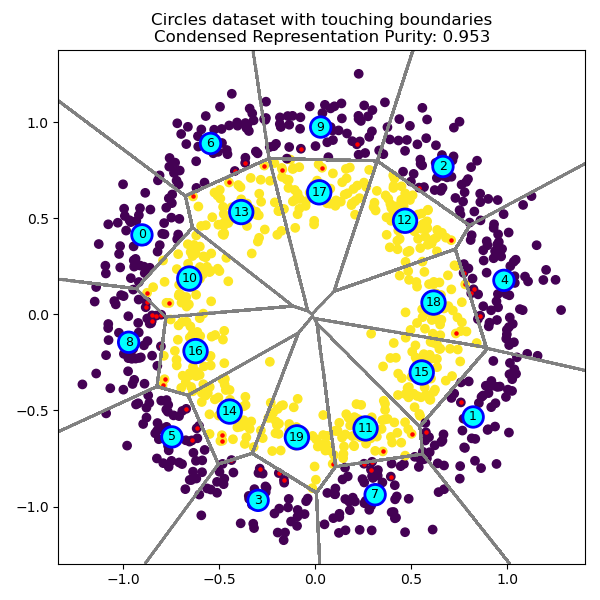}
        \caption{Touching boundaries, condensed}
        \label{fig:test_circles:2}
    \end{subfigure}
    \begin{subfigure}[b]{0.32\textwidth}
        \centering
        \includegraphics[width=\textwidth]{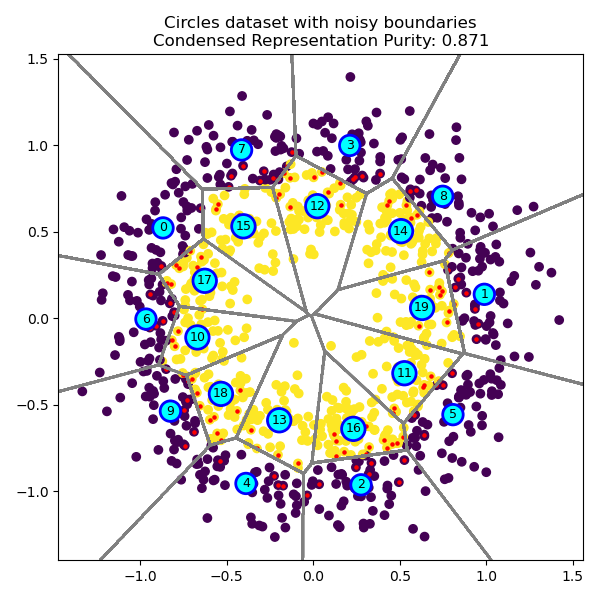}
        \caption{Noisy boundaries, condensed}
        \label{fig:test_circles:3}
    \end{subfigure}
    
    \vspace{0.1cm}

    \begin{subfigure}[b]{0.32\textwidth}
        \centering
        \includegraphics[width=\textwidth]{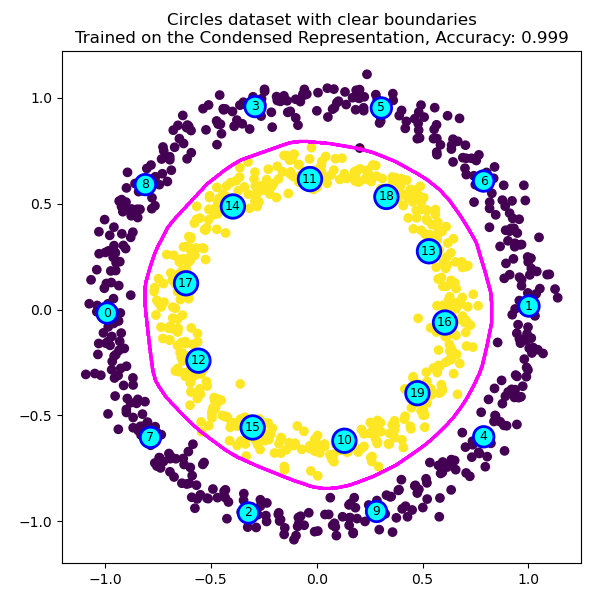}
        \caption{Clear boundaries, trained}
        \label{fig:test_circles:4}
    \end{subfigure}
    \begin{subfigure}[b]{0.32\textwidth}
        \centering
        \includegraphics[width=\textwidth]{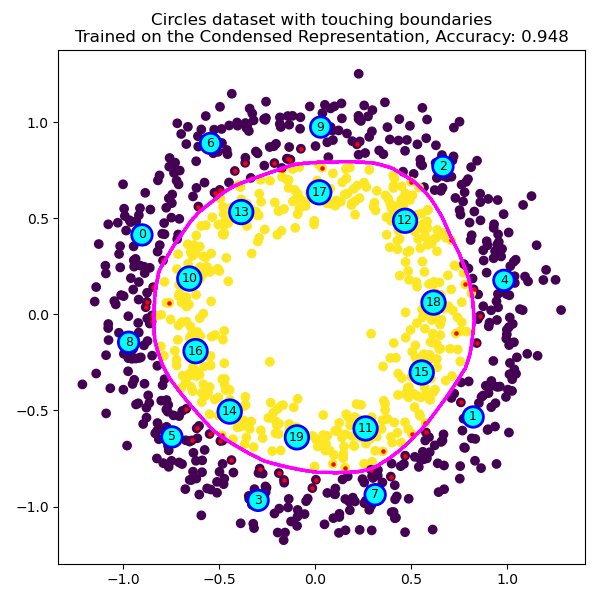}
        \caption{Touching boundaries, trained}
        \label{fig:test_circles:5}
    \end{subfigure}
    \begin{subfigure}[b]{0.32\textwidth}
        \centering
        \includegraphics[width=\textwidth]{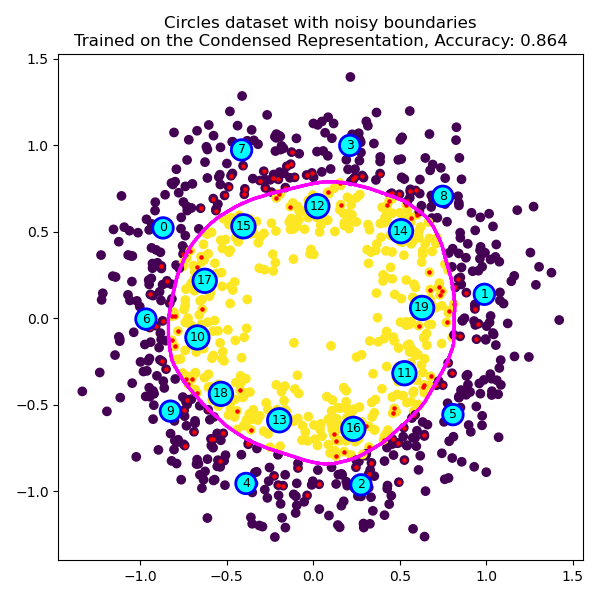}
        \caption{Noisy boundaries, trained}
        \label{fig:test_circles:6}
    \end{subfigure}
    \caption{Condensed representations of circles dataset and results of models trained on condensed representations}
    \label{fig:test_circles}
\end{figure*}

\begin{figure*}
    \centering
    \begin{subfigure}[b]{0.32\textwidth}
        \centering
        \includegraphics[width=\textwidth]{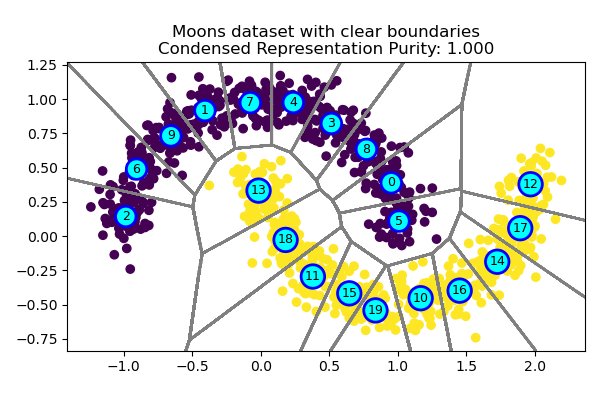}
        \caption{Clear boundaries, condensed}
        \label{fig:test_moons:1}
    \end{subfigure}
    \begin{subfigure}[b]{0.32\textwidth}
        \centering
        \includegraphics[width=\textwidth]{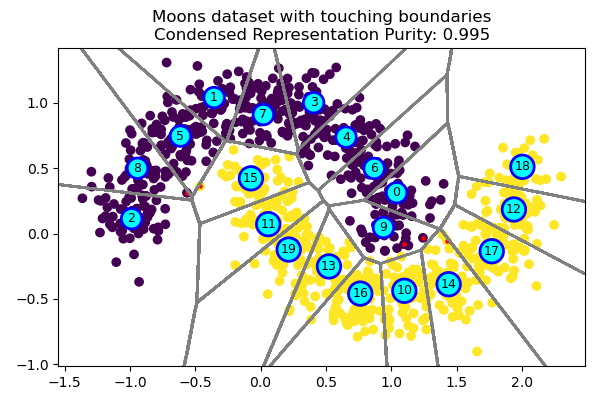}
        \caption{Touching boundaries, condensed}
        \label{fig:test_moons:2}
    \end{subfigure}
    \begin{subfigure}[b]{0.32\textwidth}
        \centering
        \includegraphics[width=\textwidth]{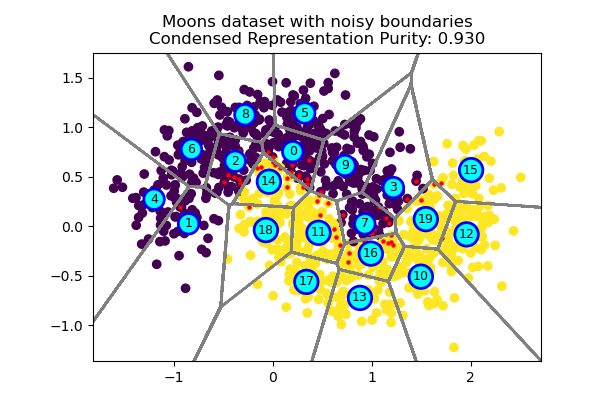}
        \caption{Noisy boundaries, condensed}
        \label{fig:test_moons:3}
    \end{subfigure}
    
    \vspace{0.1cm}

    \begin{subfigure}[b]{0.32\textwidth}
        \centering
        \includegraphics[width=\textwidth]{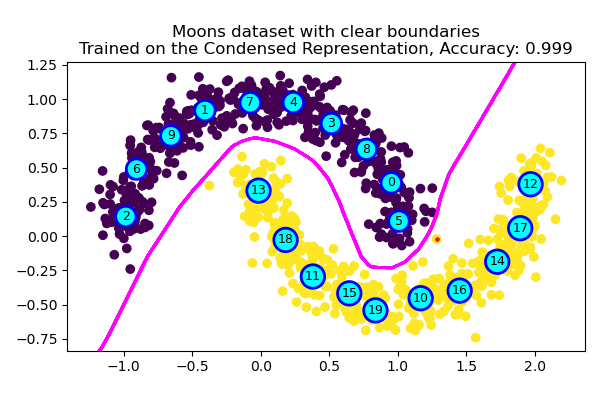}
        \caption{Clear boundaries, trained}
        \label{fig:test_moons:4}
    \end{subfigure}
    \begin{subfigure}[b]{0.32\textwidth}
        \centering
        \includegraphics[width=\textwidth]{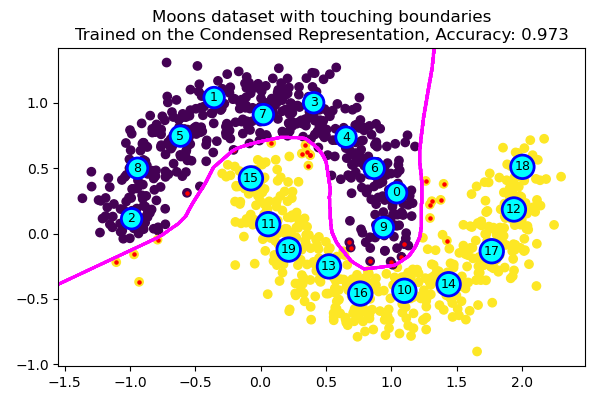}
        \caption{Touching boundaries, trained}
        \label{fig:test_moons:5}
    \end{subfigure}
    \begin{subfigure}[b]{0.32\textwidth}
        \centering
        \includegraphics[width=\textwidth]{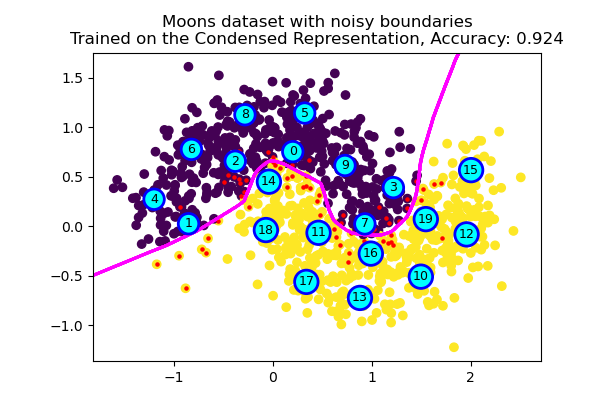}
        \caption{Noisy boundaries, trained}
        \label{fig:test_moons:6}
    \end{subfigure}
    \caption{Condensed representations of moons dataset and results of models trained on condensed representations}
    \label{fig:test_moons}
\end{figure*}

Circles and moons datasets are generated with varying noise levels using the
Scikit-Learn \cite{pedregosa_scikit-learn_2011} library functions. The
condensation is done on GPU. Using the condensed representations,
multilayer-perceptron classifier \cite{pedregosa_scikit-learn_2011} models are
trained and tested on the CPU for accuracy on the original datasets. It can be
seen in Figure \ref{fig:test_circles}) and (Figure \ref{fig:test_moons}) that
training nonlinear machine learning models on the condensed representations of
training datasets gives acceptable accuracies with the possibility of easier
training. Training the multilayer perceptron classifier with condensed
representations results in practically acceptable accuracies.

\section{Conclusion}
\label{sec:conclusion}

Massive amounts of data are continuously generated throughout many processes.
Such big data mostly consists of redundancy, so that can be reduced to a
condensed representation without losing much accuracy, especially for machine
learning applications. In this paper, an alternate way of handling machine
learning datasets is introduced. The ideas introduced in this paper can be
summarized as follows:

\begin{itemize}
    \item K-means clustering is used as a piecewise-linear classifier model
    generator.
    \item The generated PWL classifier models are composed of condensed
    representation centers, which have the same dimensionality as the data.
    \item The CRC locations can be fine-tuned for class separation purity and
    this is equivalent to improving the accuracy of a NN classifier, classifying
    on CRCs.
    \item The final CRCs can be used as training data for various nonlinear
    classifiers, such as multilayer neural networks.
\end{itemize}

The usage of condensed representation of data instead of the raw data allows
focusing on different stages of the machine-learning pipeline from different
aspects. For example, an online machine-learning pipeline can be separated into
two main stages: Online Condensation and Machine Learning. A more specific
example can be, an autonomous mobility platform that has data condensation edge
devices with central machine learning processing.

This paper is flexibly and modularly constructed, so that any sub-element of the
novel method, introduced in this paper, can be replaced or modified to study
different aspects of condensed representation.

\section*{Acknowledgments}
\label{sec:acknowledgments}

This research was supported by the Turkish Scientific and Technological Research
Council (TUBITAK) under project no. 121E537.

We would like to thank the reviewers for their thoughtful comments and their
constructive remarks.


\bibliographystyle{elsarticle-num} 
\bibliography{CRofMLD.bib}

\end{document}